\documentclass{article}

\usepackage[preprint, nonatbib]{nips_2018}
\usepackage{multirow}
\usepackage{graphicx}
\usepackage{amsmath,amssymb} 

\usepackage[square,numbers]{natbib}
\usepackage[utf8]{inputenc} 
\usepackage[T1]{fontenc}    
\usepackage{hyperref}       
\usepackage{url}            
\usepackage{booktabs}       
\usepackage{amsfonts}       
\usepackage{nicefrac}       
\usepackage{microtype}      

\title{Select, Attend, and Transfer: Light, Learnable Skip Connections}

\author{
$\text{Saeid Asgari Taghanaki}^{1,2}$, $\text{Aicha Bentaieb}^{1}$, $\text{Anmol Sharma}^{1}$, $\text{S. Kevin Zhou}^{2}$, \\ \textbf{$\text{Yefeng Zheng}^{2}$}, \textbf{$\text{Bogdan Georgescu}^{2}$}, \textbf{$\text{Puneet Sharma}^{2}$}, \textbf{$\text{Sasa Grbic}^{2}$}, \textbf{$\text{Zhoubing Xu}^{2}$}, \\ \textbf{$\text{Dorin Comaniciu}^{2}$}, \textbf{$\text{Ghassan Hamarneh}^{1}$} \\
    $\scriptsize{1} _\text{\small{Medical Image Analysis Lab, School of Computing Science, Simon Fraser University, Canada}}$ \\
    $\scriptsize{2} _\text{\small{Medical Imaging Technologies, Siemens Healthineers, Princeton, NJ, USA}}$
}

\begin{document}

\maketitle

\begin{abstract}
  Skip connections in deep networks have improved both segmentation and classification performance by facilitating the training of deeper network architectures, and reducing the risks for vanishing gradients. They equip encoder-decoder-like networks with richer feature representations, but at the cost of higher memory usage, computation, and possibly resulting in transferring non-discriminative feature maps. In this paper, we focus on improving  skip connections used in segmentation networks (e.g., U-Net, V-Net, and The One Hundred Layers Tiramisu (DensNet) architectures). We propose light, learnable skip connections which learn to first select the most discriminative channels and then attend to the most discriminative regions of the selected feature maps. The output of the proposed skip connections is a unique feature map which not only reduces the memory usage and network parameters to a high extent, but also improves segmentation accuracy. We evaluate the proposed method on three different 2D and volumetric datasets and demonstrate that the proposed light, learnable skip connections can outperform the traditional heavy skip connections in terms of segmentation accuracy, memory usage, and number of network parameters.
\end{abstract}

\section{Introduction}

Recent works in image segmentation has shown that deep segmentation networks (stacking tens of convolutional layers) generally perform better than shallower networks~\cite{he2016deep}, due to their ability to learn more complex nonlinear functions. However, they are difficult to train because of the high number of parameters and gradient vanishing. One of the recent key ideas to effectuate the training process and handle gradient vanishing is to introduce skip connections between subsequent layers in the network, which has been shown to improve some of the encoder-decoder segmentation networks (e.g., 2D U-Net~\cite{ronneberger2015u}, 3D U-Net~\cite{cciccek20163d}, 3D V-Net~\cite{milletari2016v}, and The One Hundred Layers Tiramisu (DensNet)~\cite{jegou2017one}). Skip connections help in the training process by recovering spatial information lost during down-sampling, as well as reducing the risks of vanishing gradients~\cite{huang2017densely}. It is also has been shown that the skip connections eliminate singularities~\cite{orhan2017skip}. However, direct transfer of feature maps from previous layers to subsequent layers may also lead to redundant and non-discriminatory feature maps being transferred. Also, as the transferred feature maps are concatenated to the feature maps in subsequent layers, the memory usage increases many folds.

\textbf{Complexity reduction.} Recently, there have been several efforts to reduce the training and runtime computations of deep classification networks~\cite{leroux2018iamnn,howard2017mobilenets, zhang2017shufflenet, iandola2016squeezenet}. A few other works have attempted to simplify the structure of deep networks, e.g., by tensor factorization~\cite{jaderberg2014speeding,kim2015compression,lebedev2014speeding}, channel/network pruning~\cite{wen2016learning,hu2016network} or applying sparsity to connections~\cite{han2015learning,han2015deep,liu2015sparse,guo2016dynamic,han2016eie}. However, the non-structured connectivity and irregular memory access, which is caused by sparsity regularization and connection pruning methods, adversely impacts practical speedup~\cite{wen2016learning}. On the other hand, tensor factorization is not compatible with the recently designed networks, e.g., GoogleNet and ResNet, and many such methods may even end up with more computations than the original architectures~\cite{he2017channel}. Wei et al.~\cite{wen2016learning}, introduced a learning based sparsity approach by adding sparsity terms in their optimization function and leveraging group Lasso~\cite{yuan2006model}. Similarly. Alvarez et. al.~\cite{alvarez2016learning}, added a regularization term to their loss function to reduce the number of neurons in a learnable setting. However, for both works, it is not trivial to decide on the level of contribution of each term in their loss function. Hu et al.~\cite{hu2016network} proposed a neuron pruning method for having an efficient architecture. Although their method performs at par with the original unpruned model, it is a complex multi-stage threshold based method applied only to parameter-dense layers of the network. Yang et al.~\cite{yang2017designing} proposed an energy-aware method to reduce the hardware energy consumption by pruning convolutional neural networks. However, similar to~\cite{liu2015sparse,denton2014exploiting,jaderberg2014speeding,he2017channel}, their approach traded off network performance  for reduced computational complexity/energy consumption (i.e., they reported lower accuracy scores along with   lower energy consumption), which may not be desirable in some applications, especially in medical image segmentation. 

\textbf{Gates and attention.} Attention can be viewed as using information transferred from several subsequent layers/feature maps to localize the most discriminative (or salient) part of the input signal. Attention models have been widely applied for machine translation~\cite{bahdanau2014neural}, visual question answering~\cite{das2017human}, sequence based models~\cite{luong2015effective}, and image captioning~\cite{xu2015show}. Srivastava et al.~\cite{srivastava2015highway} modified ResNet in a way to control the flow of information through a connection; their proposed gates control the level of contribution between unmodified input and activations to a consecutive layer. Hu et al.~\cite{hu2017squeeze} proposed a selection mechanism where feature maps are first aggregated using global average pooling and reduced to a single channel descriptor, then an activation gate is used to highlight the most discriminative features. Recently, Wang et al.~\cite{wang2017residual} added an attention module to ResNet for image classification. Their proposed attention module consists of several encoding-decoding layers, which although helped in improving image classification accuracy, also increased the computational complexity of the model by an order of magnitude~\cite{wang2017residual}.

\begin{figure}
\centering
\includegraphics[scale=0.6]{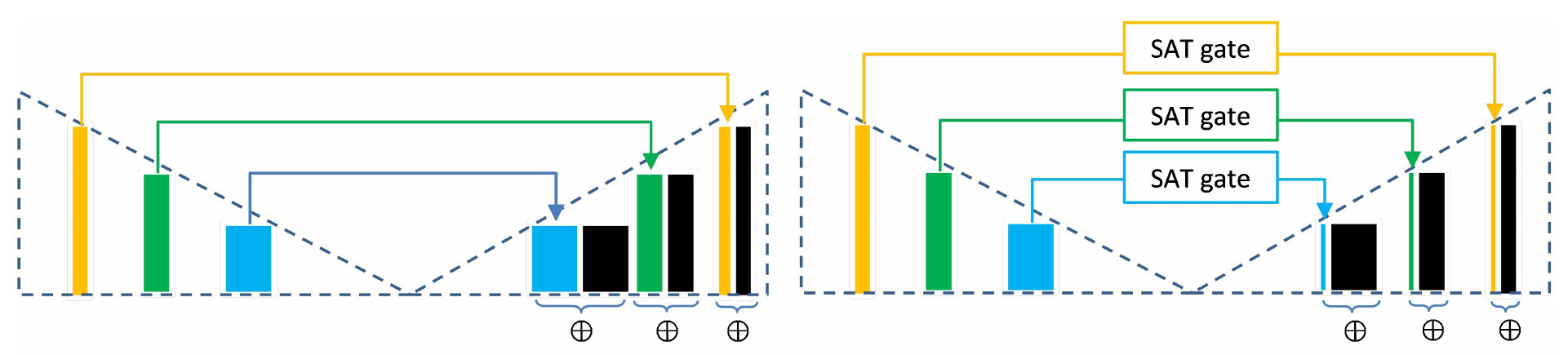}
\caption {We propose light, learnable skip connections that improve segmentation accuracy while transferring only one channel instead of many from encoder to decoder. The SAT gate refers to the proposed select-attend-transfer gate. The $\oplus$ symbol denotes concatenation and black bars show decoder blocks composed of convolution, activation function, and up-sampling. Left: traditional skip connections; right: proposed skip connections.}
\label{figure1}
\end{figure}

In this paper, we propose a modification of the traditional skip connections, using a novel \emph{select-attend-transfer gate}, which aims at simultaneously improving segmentation accuracy and reducing memory usage and network parameters (Fig.~\ref{figure1}). We focus on skip connections in encoder-decoder architectures (i.e. as opposed to skip connections in residual networks) designed for segmentation tasks. Our proposed \emph{select-attend-transfer gate} favours sparse feature representations and uses this property to select and attend to the most discriminative feature channels and spatial regions within a skip connection. Specifically, we first learn to identify the most discriminative feature maps in a skip connection, using a set of trainable weights under sparsity constraint. Then, we reduce the feature maps to a single channel using a convolutional filter followed by an attention layer to identify salient spatial locations within the produced feature map. This compact representation forms the final feature map of the skip connection. 

Note that our feature selection method differs from these previous works, most notably because: a) instead of indirectly training for selection, i.e., by incorporating new terms in the objective function~\cite{wen2016learning}, we directly train the selection parameters along with other parameters of the network; b) in contrast to previous works mentioned above, we focus specifically on improving segmentation accuracy and consistency by re-designing the skip connections within fully convolutional encoder-decoder networks; and c) instead of transferring all the channels, the proposed method transfers only one attention map, which reduces memory usage and network parameters.

\section{The Select-Attend-Transfer (SAT) gate}

\textbf{Notation:} We define $f$ as an input feature map of size $H\times W\times D\times C$ where $H$, $W$, $D$, and $C$ refer to the height, width, and depth of the volumetric data, and number of channels, respectively. The notation in the paper is defined for 3D (volumetric) input images but the method can be easily adapted to 2D images by removing an extra dimension and applying 2D convolutions instead of 3D.

\begin{figure}
\centering
\includegraphics[scale=.45]{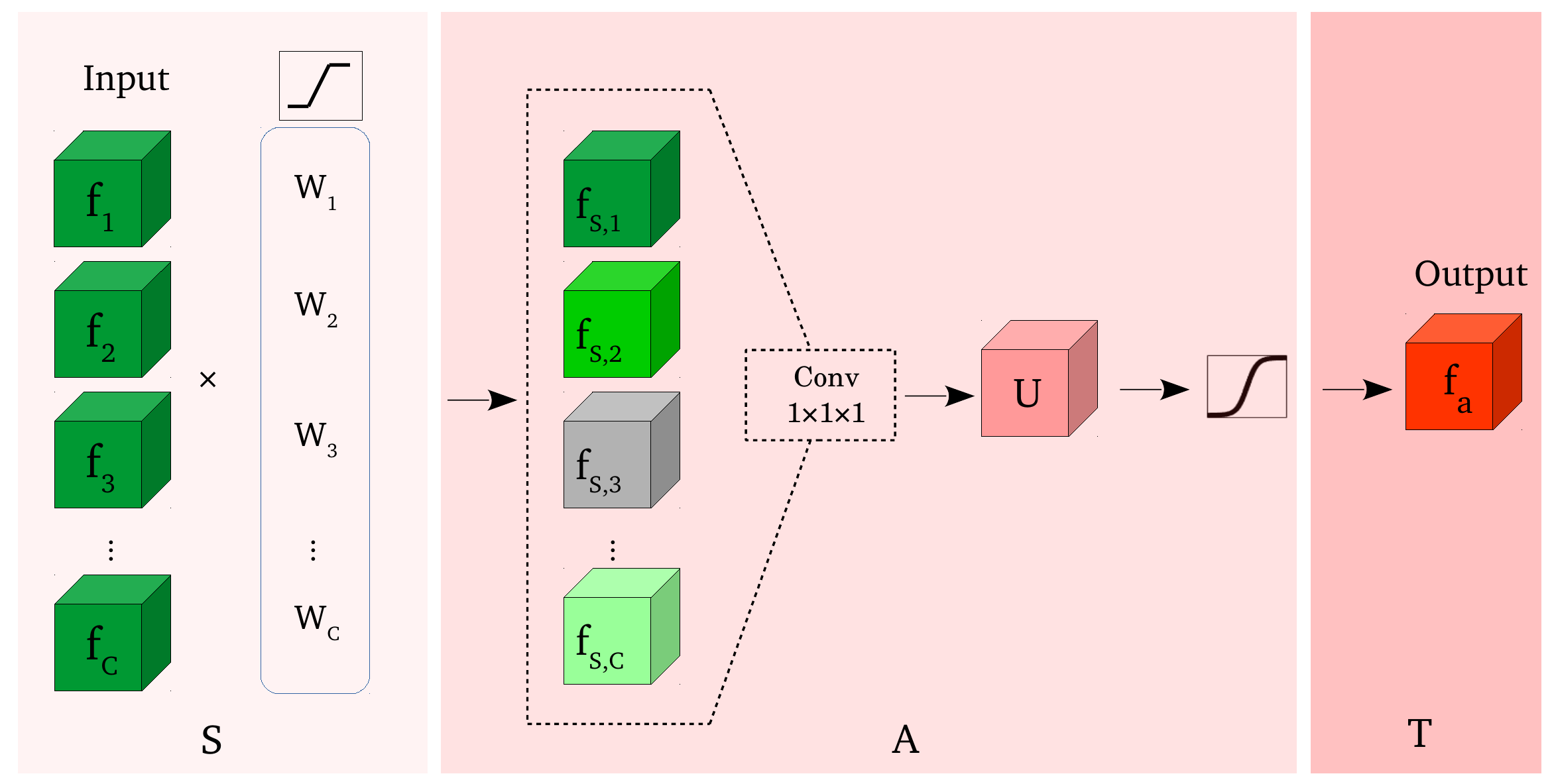}
\caption{The proposed select-attend-transfer (SAT) gate for a volumetric 3D data. The different shades of green colour show the result of applying the proposed soft channel selection, i.e., some channels can be totally turned off (grey) or partially turned on.}
\label{fig2}
\end{figure}

\noindent An overview of the proposed SAT gate is shown in Fig.~\ref{fig2}. It consists of the following modules: 1) Select: re-weighting the channels of the input feature maps $f$, using a learned weight vector with sparsity constraints, to encourage sparse feature map representations, that is, only those channels with non-zero weights are selected; 2) Attend: discovering the most salient spatial locations within the final feature map; and 3) Transfer: transferring the output of the gate into subsequent layers via a skip connection.

\noindent \textbf{Selection via sparsification.} We weight the channels of an input feature map $f$ by using a scalar weight vector $W$ trained along with all other network parameters. The weight vector is defined such that we encourage sparse feature maps, resulting in completely/partially turning off or on feature maps. Instead of relying on complex loss functions, we clip negative weights to zero and positives to at most one using a truncated ReLU function~\cite{lecun2015deep}. Each channel of the feature map block $f$ is multiplied by $W$ as the following (eq. 1)

\begin{equation}
f_{s,t} = f_{t}\ast trelu (W_t)
\end{equation}

\begin{equation}
trelu(z) = \{z \ \textrm{if} \ z\in [0,1]; \ 0 \ \textrm{if} \ z<0; \ 1 \ \textrm{if} \ z>1. \}
\end{equation}

\noindent where $W_t$ is a scalar weight value associated with the input feature map channel $f_t$; $trelu(.)$ is the truncated ReLU function. The output $f_s$ is of size $H\times W\times D\times C$ and is a sparse representation of $f$. Zero weights turn off corresponding feature maps completely, whereas positive weights fully/partially turn on features maps; i.e., implementing the soft feature map selection.

\noindent \textbf{Attention via filtering.} The output $f_s$ is further filtered to identify the most discriminative linear combination of feature channels $C$. For this, we employ a unique convolution filter $K$ of size $1\times 1\times 1\times C$ ($1\times 1\times C$ for 2D data), which allows us to learn how best to aggregate the different channels of the feature map $f_s$. The output (i.e. $U$) of this feature selection step reduces $f_s$ to $K\star f_s$ of size $H\times W\times D\times 1$ (eq. 3). 

\begin{equation}
U=(K\star f_s)
\end{equation}


\noindent where $\star$ is the convolution operation. To identify salient spatial locations within the $U$, we introduce an attention gate $f_a$ (eq. 4) composed of a sigmoid activation function. Using the sigmoid as an activation function allows us to identify multiple discriminative spatial locations (as opposed to one single location, which is the case with softmax) within the feature map $U$.

\begin{equation}
f_a= \sigma(U)
\end{equation}

\noindent where $\sigma$ denotes the sigmoid function. The computed $f_a$ forms a compact summary of the input feature map $f$.

\noindent \textbf{Transfer via skip connection.} The computed $f_a$ is transferred to subsequent layers via a skip connection.

\noindent \textbf{Special examples.} There are two special cases of the proposed SAT gate. One is the ST gate which skips the A part, that is, only channel selection but no attention is performed. The signal $f_s$ is directly fed to subsequent layers. The other is the AT gate which skips the S part by setting all weights to one. This way, there is no channel selection and only attention is performed.

\noindent \textbf{Training and implementation details.} Networks are trained using stochastic gradient descent with momentum. We do not rely on special layers or non-differentiable operations to permit model training via standard backpropagation. To set the hyperparameters, we started with the proposed values mentioned in U-Net, V-Net, and The One Hundred Layers Tiramisu papers. However, we found experimentally that applying ADADELTA~\cite{zeiler2012adadelta} optimizer (with its proposed default parameters: $learning \ rate=1$, $\rho=0.95$, $\epsilon=1e-8$, and $decay=0$) with Glorot uniform model initializer, also called Xavier uniform initializer~\cite{glorot2010understanding}, works best for all the networks. All the models are implemented using Keras with TensorFlow backend. After each convolution layer we use batch normalization. This allows us to use higher learning rate since the effect of outliers are reduced by batch normalization. Note that to test on 2D dataset, we replace all the 3D opperations in 3D-Vnet with 2D operations and for 3D datasets we replace all the 2D operations of The One Hundred Layers Tiramisu model with 3D ones.

\section{Experiments}
\begin{figure}
\centering
\includegraphics[scale=.35]{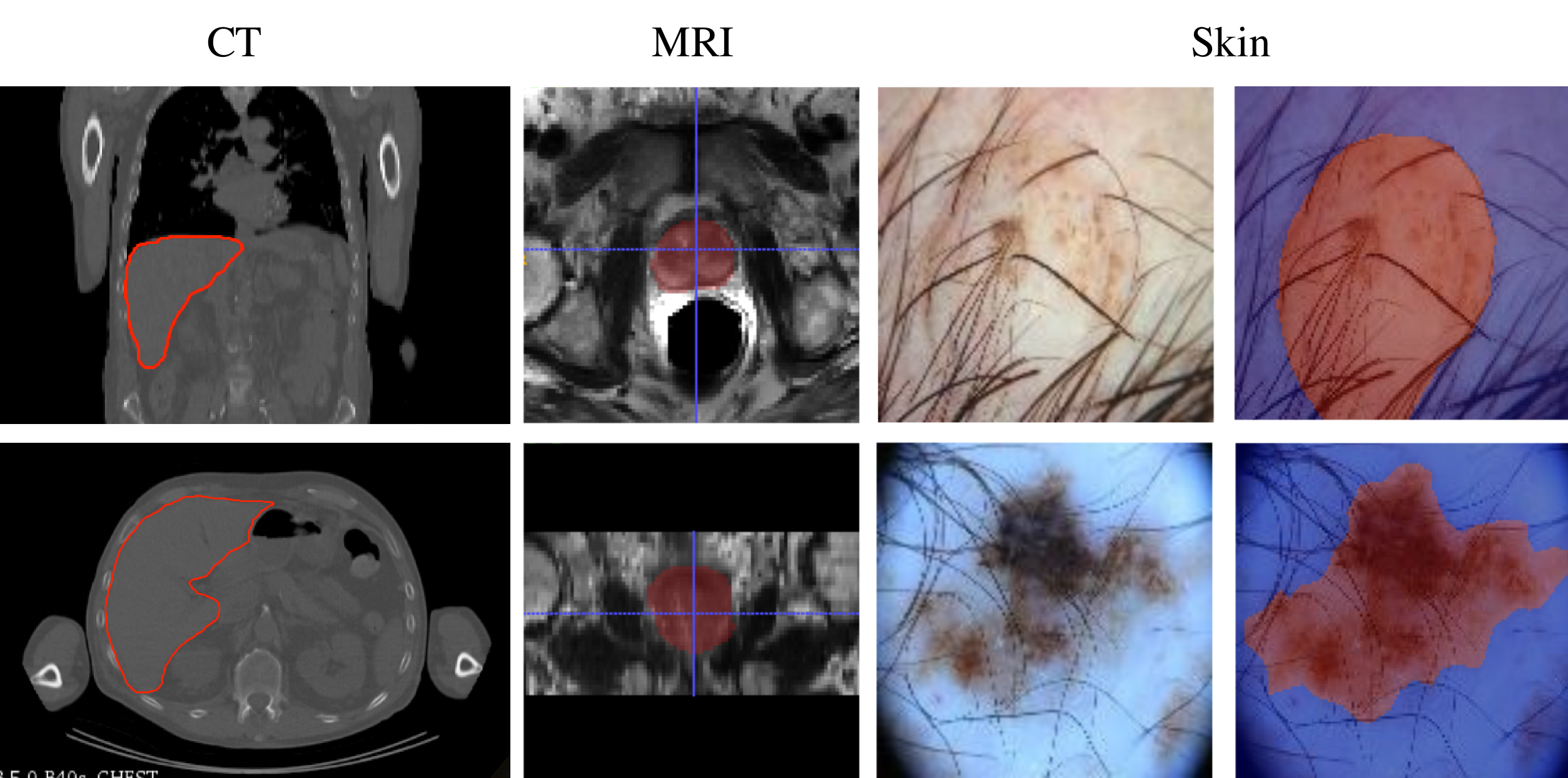}
\caption{Several samples of the used datasets. The first column shows the liver (contoured in red) in CT scan. In the second column, the prostate is highlighted in red in an MRI scan. For CT and MRI samples, the 2 rows show the axial and coronal views. Column 3  shows 2 sample skin images containing lesions, which highlighted in red in column 4.}
\label{figure3}
\end{figure}

In this section, we evaluate the performance of the proposed method on three commonly used fully convolutional segmentation networks which leverage skip connections: U-Net (both 2D and 3D), V-Net (both 2D and 3D), and the One Hundred Layers Tiramisu network (both 2D and 3D). As U-Net and V-Net were originally designed for biomedical image segmentation, we tested the proposed method on a series of volumetric (3D) and 2D medical imaging datasets (Fig.~\ref{figure3}) including (i) a magnetic resonance imaging (MRI) dataset; (ii) a skin lesion dataset; and (iii) a computed tomography (CT) dataset.

To analyze the performance of the proposed method, we performed the following experiments: a) We tested the performance of the proposed method, in terms of segmentation accuracy, on datasets i and ii.
b) We applied the proposed SAT gate to the recently introduced method Deep Image to Image Network (DI2IN) ~\cite{yang2017automatic} for liver segmentation method (Section 3.2).
c) We quantitatively and qualitatively analysed the proposed skip connections in terms of the amount of data transferred and we visualized the outputs of both channel selection and attention layers. We also compared the proposed method vs. the original networks in terms of memory usage and number of parameters (Section 3.3). Note that for the skin 2D dataset, as original V-Net is 3D, we replaced all the 3D operations in V-Net with 2D counterparts, and for testing the One Hundred Layers Tirsamisu model with volumetric data, we replaced all the network's 2D operations with 3D ones.

\subsection{Volumetric and 2D binary segmentation}

\textbf{Volumetric CT liver segmentation.} In this experiment, the goal is to segment the liver from CT images. We used more than 2000 CT scans (to the best of our knowledge this is the largest CT dataset used in the literature so far) of different resolutions collected internally and from The Cancer Imaging Archive (TCIA) QIN-HEADNECK dataset~\cite{QINheadNeck2015, QINheadNeck2016, clark2013cancer}, which were resampled to isotropic voxels of size $2\times2\times2 \ (mm)$. The QIN-HEADNECK dataset is originally collected from a set of head and neck cancer patients. It has multiple whole body positron emission tomography/computed tomography (PET/CT) scans before and after therapy. However, here we use the CT scans for the purpose of liver segmentation. We picked 61 challenging volumes of whole dataset for testing and trained the networks on the remaining volumes.

\noindent \textbf{Volumetric MRI prostate segmentation.} In this experiment, we test the proposed method on a volumetric MRI prostate dataset. The dataset contains 1029 MRI volumes of different resolutions which were collected internally and from TCIA ProstateX dataset~\cite{ProstateX2017,litjens2014computer,clark2013cancer} that were resampled to isotropic voxels of size $2\times2\times2 \ (mm)$. We used 770 images for training and 258 images for testing. 

\noindent \textbf{2D RGB skin lesion segmentation.} For this experiment, we used the 2D RGB skin lesion dataset from the 2017 IEEE ISBI International Skin Imaging Collaboration (ISIC) Challenge~\cite{codella2017skin}. We train on a dataset of 2000 images and test on a different set of 150 images.

As reported in Table~\ref{table1}, overall, equipping U-Net and V-Net with SAT improved Dice results for all  4 (2 modalities times 2 networks) experiments. Specifically: (i) For the MRI dataset, the SAT gate improves U-Net performance by 1.15\% (0.87 to 0.88) and 35.3\% (0.17 to 0.11) in Dice and FNR, respectively. Using SAT with V-Net improved results by 2.4\% (0.85 to 0.87) and 20\% (0.005 to 0.004) in Dice and FPR. Note that, although for V-Net (MRI data) the Dice improvement is small, instead of transferring all the channels, the proposed method transfers only one attention map, which reduces memory usage to a high extent (4th column in Table 28). (ii) For the skin dataset, equipping U-Net with SAT improved Dice and FPR by 2.53\% (0.79 to 0.81) and 25\% (0.04 to 0.03), and similarly V-Net with SAT resulted in 2.5\% (0.81 to 0.83) and 35\% (0.20 to 0.13) improvement in terms of Dice and FNR, respectively.

Note also that, although sometimes ST obtained almost similar results to SAT, the SAT transfers only one channel whereas ST results in transferring multiple channels (i.e. ST requires more parameters and involves higher memory usage). Similar to SAT, AT also transfers only one channel, however, as shown in Figures 6 and 7, AT tends to attend to wrong regions as it does not leverage the ST module to filter the most discriminative channels.

\begin{table}
\small
\setlength{\tabcolsep}{5pt}
\centering
\caption{2D/3D U-Net and 2D/3D V-Net results of prostate segmentation from volumetric MRI images and skin lesion segmentation. ORG, AT, CS and SAT refer to the original skip connections which transfer all the data (i.e ORG), skip connections which transfer only an attention map (i.e. AT) skip connections which transfer several soft selected channels (i.e. ST), and skip connections which transfer only one attention map produced from soft selected channels (i.e. SAT). N is the total number of channels C before skip connection.}
\label{table1}
\begin{tabular}{lcllccc}
\hline
Data & \multicolumn{1}{l}{Model} & Method & \#C & Dice & FPR & FNR \\ \hline

\multirow{8}{*}{MRI} & \multirow{4}{*}{3D U-Net~\cite{cciccek20163d}} & ORG & C = N & $0.87\pm0.05$ & $0.002\pm0.004$ & $0.17\pm0.09$ \\
 &  & AT & C = 1 & $0.88\pm0.05$ & $0.004\pm0.003$ & $0.09\pm0.09$ \\
 &  & ST & $\text{C} \leqslant \text{N}$ & $0.88\pm0.05$ & $0.003\pm0.003$ & $0.12\pm0.09$ \\
 &  & SAT & C = 1 & $0.88\pm0.05$ & $0.003\pm0.004$ & $0.11\pm0.09$ \\ \cline{2-7} 
 & \multirow{4}{*}{3D V-Net\cite{milletari2016v}} & ORG & C = N & $0.85\pm0.05$ & $0.005\pm0.004$ & $0.09\pm0.07$ \\
 &  & AT & C = 1 & $0.86\pm0.05$ & $0.004\pm0.004$ & $0.12\pm0.07$ \\
 &  & ST & $\text{C} \leqslant \text{N}$ & $0.86\pm0.05$ & $0.005\pm0.003$ & $0.09\pm0.07$ \\
 &  & SAT & C = 1 & $0.87\pm0.04$ & $0.004\pm0.003$ & $0.11\pm0.07$ \\ \hline
\multirow{8}{*}{Skin} & \multicolumn{1}{l}{\multirow{4}{*}{2D U-Net~\cite{ronneberger2015u}}} & ORG & C = N & \multicolumn{1}{l}{$0.79\pm0.22$} & \multicolumn{1}{l}{$0.040\pm0.070$} & \multicolumn{1}{l}{$0.16\pm0.22$} \\
 & \multicolumn{1}{l}{} & AT & C = 1 & \multicolumn{1}{l}{$0.79\pm0.20$} & \multicolumn{1}{l}{$0.030\pm0.050$} & \multicolumn{1}{l}{$0.20\pm0.22$} \\
 & \multicolumn{1}{l}{} & ST & $\text{C} \leqslant \text{N}$ & \multicolumn{1}{l}{$0.79\pm0.22$} & \multicolumn{1}{l}{$0.020\pm0.040$} & \multicolumn{1}{l}{$0.22\pm0.24$} \\
 & \multicolumn{1}{l}{} & SAT & C = 1 & \multicolumn{1}{l}{$0.81\pm0.20$} & \multicolumn{1}{l}{$0.030\pm0.050$} & \multicolumn{1}{l}{$0.18\pm0.23$} \\ \cline{2-7} 
 & \multicolumn{1}{l}{\multirow{4}{*}{2D V-Net}} & ORG & C = N & \multicolumn{1}{l}{$0.81\pm0.21$} & \multicolumn{1}{l}{$0.020\pm0.040$} & \multicolumn{1}{l}{$0.20\pm0.24$} \\
 & \multicolumn{1}{l}{} & AT & C = 1 & \multicolumn{1}{l}{$0.82\pm0.23$} & \multicolumn{1}{l}{$0.020\pm0.040$} & \multicolumn{1}{l}{$0.16\pm0.25$} \\
 & \multicolumn{1}{l}{} & ST & $\text{C} \leqslant \text{N}$ & \multicolumn{1}{l}{$0.82\pm0.22$} & \multicolumn{1}{l}{$0.020\pm0.040$} & \multicolumn{1}{l}{$0.18\pm0.22$} \\
 & \multicolumn{1}{l}{} & SAT & C = 1 & \multicolumn{1}{l}{$0.83\pm0.20$} & \multicolumn{1}{l}{$0.030\pm0.050$} & \multicolumn{1}{l}{$0.13\pm0.21$} \\ \hline
\end{tabular}
\end{table}

\begin{figure}[h]
\centering
\includegraphics[scale=0.5]{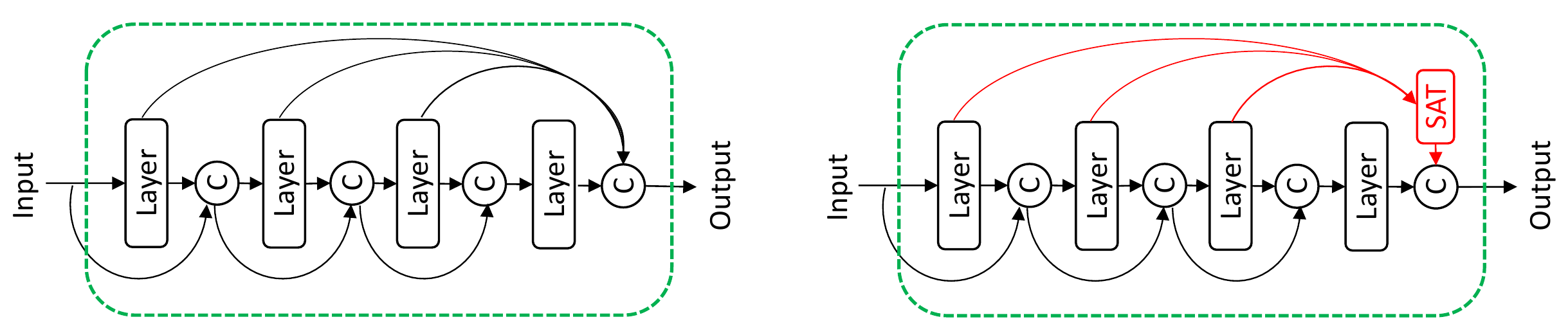}
\caption {Placing a SAT gate inside of a dense block; left: original; right: proposed. C in the figure refers to concatenation operation.}
\label{densblock}
\end{figure}

We further test the proposed SAT gate on The One Hundred Layers Tiramisu network~\cite{jegou2017one}. Note that we apply the proposed SAT gate to more skip connections in this network i.e. both long skip connections between the encoder and decoder (similar to Figure~\ref{figure1}) and the long skip connections inside of each dense block (Figure~\ref{densblock}). While our proposed SAT gate reduces the number of parameters in the original Tiramisu network (Table~\ref{table3}) by $\sim20.7\%$, it improves Dice results (Table~\ref{tiramisu}) by 3.7\% and 2.5\% for MRI and skin datasets, respectively.

\begin{table}[h]
\small
\setlength{\tabcolsep}{6pt}
\centering
\caption{Segmentation results of The One Hundred Layers Tiramisu network on two datasets of MRI and skin before/after applying the proposed SAT gate. N is the total number of channels C before skip connection.}
\label{tiramisu}
\begin{tabular}{lllccc}
\hline
                      &     & \#C & Dice          & FPR             & FNR           \\ \hline
\multirow{2}{*}{MRI}  & ORG & C=N & $0.79\pm0.12$ & $0.006\pm0.009$ & $0.21\pm0.17$ \\
                      & SAT & C=1 & $0.82\pm0.07$ & $0.004\pm0.003$ & $0.17\pm0.13$ \\ \hline
\multirow{2}{*}{Skin} & ORG & C=N & $0.79\pm0.24$ & $0.020\pm0.050$ & $0.19\pm0.27$ \\
                      & SAT & C=1 & $0.81\pm0.20$ & $0.030\pm0.050$  & $0.15\pm0.23$ \\ \hline
\end{tabular}
\end{table}

As the next experiment, we applied our proposed SAT gate to the DI2IN~\cite{yang2017automatic} method for liver segmentation from CT images and achieved the same performance as the original DI2IN network i.e.  Dice score of 0.96 while reducing the number of parameters by 12\% (from 2,353,089 to 2,063,053) and reducing number of channels for each skip connection to only 1 channel.

\subsection{Quantitative and qualitative analysis of the proposed skip connections}

\begin{figure}
\centering
\includegraphics[scale=.7]{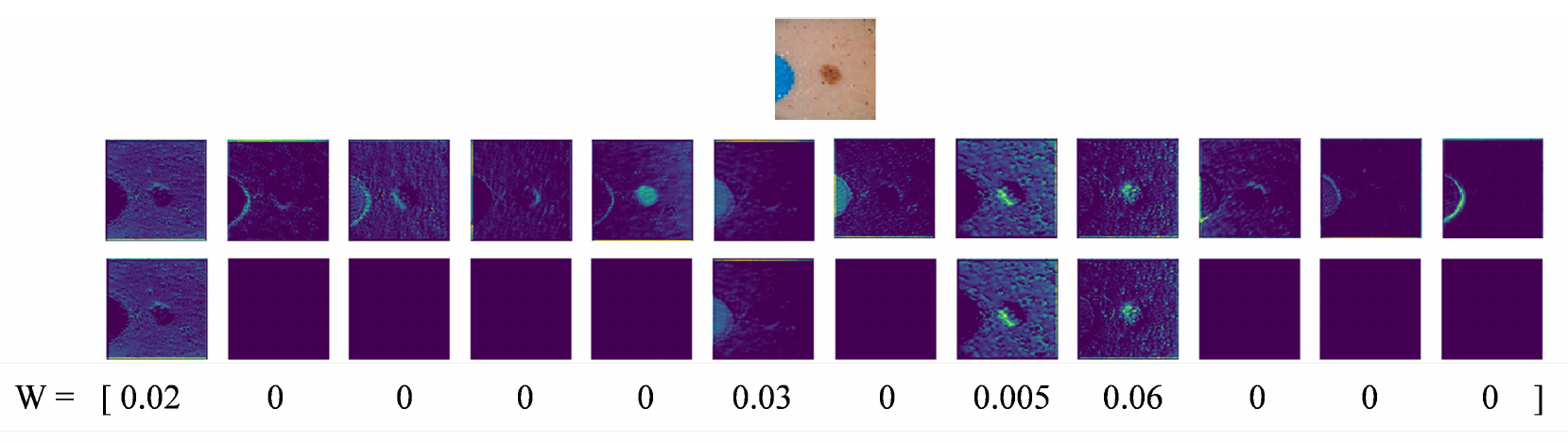}
\caption{A few samples of channel selection layer output; first row: before channel selection; second row: after channel selection. $W$ encodes the weights that have been assigned to the channels after selection.}
\label{figure5}
\end{figure}

In this section, we visualize the outputs of both the channel selection and attention layers. As shown in Fig.~\ref{figure5}, after applying the selection step, some of the less discriminative channels are completely turned off by the proposed channel selection.

\begin{figure}
\centering
\includegraphics[scale=.7]{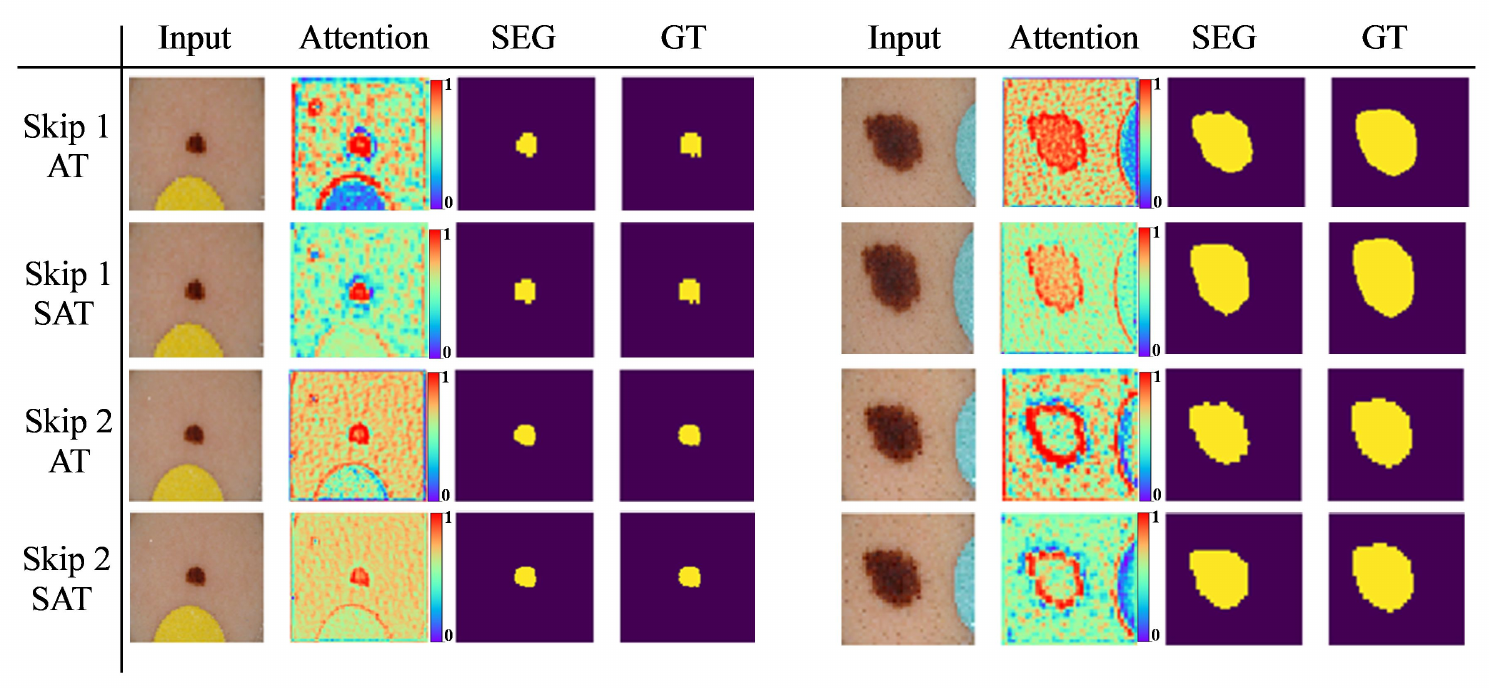}
\caption{Two samples of skin images: attention maps, segmentation, and ground truth for 2 skip connections of 2D U-Net for only attention (AT) and channel selection + attention (SAT). Left and right panels show the images corresponding to the two different samples.}
\label{figure6}
\end{figure}

\begin{figure}
\centering
\includegraphics[scale=.7]{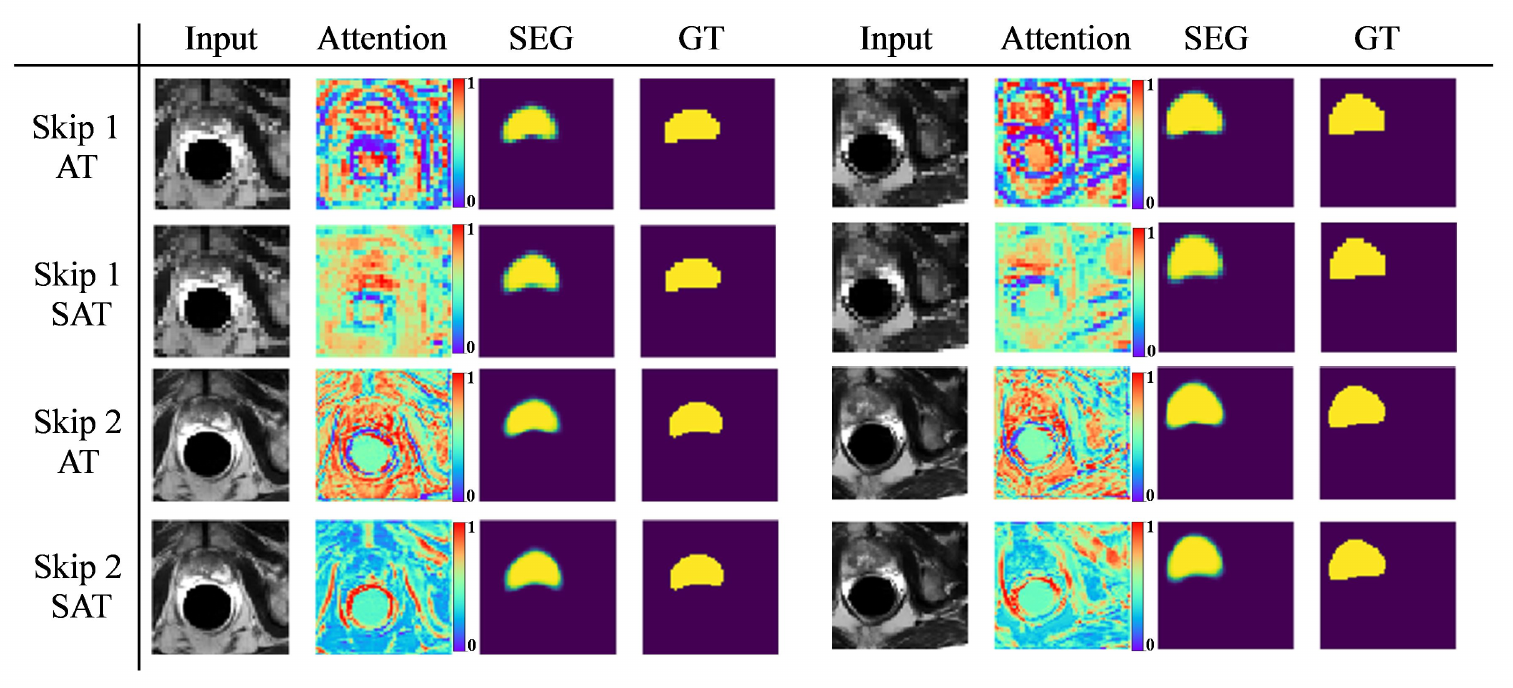}
\caption{2D views of attention volumetric prostate images, attention maps, segmentation, and ground truth for 2 skip connections of 3D V-Net for only attention (AT) and channel selection + attention (SAT). Left and right images belong to two different samples.}
\label{figure7}
\end{figure}

After the selected channels are transferred to the attention layer, the model learns to attend to the most important part(s) of the image, which helps in segmenting the target object(s) more accurately. A few samples of the attention maps for two of the proposed skip connections, used within the 2D U-Net (Fig.~\ref{figure6}) and 3D V-Net (Fig.~\ref{figure7}) architectures, are shown in Figs.~\ref{figure6} and~\ref{figure7} for the 2D skin lesion and 3D prostate MRI datasets. As can be seen in both Figs.~\ref{figure6} and~\ref{figure7}, a model with only attention layer (i.e., only AT) tends to attend to several areas of the image; including both where the object is present and absent (note the red colour visible over the whole image). However, applying channel selection (i.e., ST) before the attention layer curtails the model from attending to less discriminative regions.

We also quantitatively analyzed the proposed learnable skip connections in terms of percentage of channels “turned off” (i.e., channels i for which $w_i$ is zero in the channel selection layer). For U-Net, $55.7\pm5$ and $66.7\pm7$ and for V-Net, $57.0\pm12.9$ and $54.5\pm6.9$ percentage of channels were off for MRI and skin datasets, respectively. Since transferring only one channel (instead of N channels) as the output of the SAT gate, reduces the needed number of convolution kernel parameters in the other side of skip connections, we further report the total number of parameters before and after applying the proposed method (i.e., SAT) in Table~\ref{table3}. As can be seen, the total number of parameters are reduced by 30.2\% and 7.3\% and 30.6\% and 8.02\%, for U-Net and V-Net, respectively for MRI and Skin datasets. For The One Hundred Layers Tiramisu network the number of parameters are reduced by $\sim20.7\%$ for all the datasets. Further note that the proposed method reduces the number of convolution operations after each concatenation to almost 50\% as demonstrated next. As an example, after an original skip connection that carries 256 channels and concatenates them with another 256 channels on the other side of the network, the consequent convolution layer right after the concatenation will need 512 (=256+256) convolution operations. However, as the proposed skip connections carry only one channel, for the same example, only 257 (=1+256) convolutions are needed. Note that the reason for the difference in the number of parameters in the original networks in Table~\ref{table3} is that because of memory limitation we reduced the number of layers and/or channels for different datasets.

\begin{table}
\small
\setlength{\tabcolsep}{5pt}
\centering
\caption{Number of network parameters before and after applying the proposed method}
\label{table3}
\begin{tabular}{llccc}
\hline
& Method   & 2D/3D U-Net~\cite{ronneberger2015u,cciccek20163d}  & 2D/3D V-Net~\cite{milletari2016v} & 2D/3D Tiramisu~\cite{jegou2017one}       \\ \hline
\multirow{2}{*}{MRI}  & Original  & 2,188,289 & 76,224,513 & 10,202,993\\
                      & Proposed & 1,614,186 & 70,846,893 & 8,453,632 \\ \hline
\multirow{2}{*}{Skin} & Original & 772,401   & 13,923,041  & 3,459,185\\
                      & Proposed & 531,338   & 12,849,421 & 2,869,312\\ \hline
\end{tabular}
\end{table}

\section{Discussion}
\label{sec:blind}
\textbf{Advantages of channel selection and attention.} For some experiments, we observed slight improvement in terms of Dice score when comparing the proposed method vs. only channels selection (or attention), but we argue that there are clear advantages in combining channel selection and attention because: 1) channel selection reduces the number of feature maps and more importantly enforces learning sparse representations, 2) channel selection alone outputs many channels which require more memory and more parameters to learn whereas the proposed method transfers only one channel (which is more interpretable) thus reducing number of network parameters and memory usage, 3) attention after the selection helps focusing on the most important spatial regions in the input, which is not captured by the channel selection part, however, attention without channel selection tends to attend to wrong areas as visualized above (Figures~\ref{figure6} and~\ref{figure7}). 

\noindent \textbf{More consistent segmentations.} The proposed method by selecting most relevant features helps achieve accurate segmentations more consistently, i.e., smaller standard deviation compared to original methods (see the standard deviations in Table~\ref{table1} ).

\noindent \textbf{Using ReLu vs softmax or sigmoid in channel selection.} An advantage of ReLu is that it is computationally simpler in contrast to sigmoid which requires computing an exponent. This advantage is more obvious for deeper networks. Empirically we observed better performance using ReLu than hard sigmoid or sigmoid.

\noindent \textbf{Applying the proposed SAT gate to densely connected networks.} Dense connections make it possible to design deep architectures as they help preventing gradient vanishing. On the other hand, high memory usage (because of many concatenations) of the dense networks is a disadvantage that might limit the number of layers inside of each dense block or the number of dens blocks themselves. However, our proposed SAT gate helps reducing memory usage in densely connected networks thus using more layers/dense blocks if needed.

\section{Conclusions}

We proposed a novel architecture for skip connections in fully convolutional segmentation networks. Our proposed skip connection involves a channel selection step followed by an attention gate. Equipping popular segmentation networks (e.g. U-Net, V-Net, and The One Hundred Layers Tiramisu network) with the proposed skip connections allowed us to reduce computations and network parameters (the proposed method transfer only one unique feature channel instead of many), improve the segmentation results (it attends to the most discriminative channels and regions within the feature maps of a skip connection) and consistently obtain more accurate segmentation results. 

\noindent \textbf{Disclaimer:} This feature is based on research, and is not commercially available. Due to regulatory reasons its future availability cannot be guaranteed.

\bibliography{nips_2018}
\end{document}